\documentclass[sigconf,nonacm,natbib=true]{acmart}

\AtBeginDocument{%
}

\usepackage{amsmath,amsthm}
\usepackage{mathtools}
\usepackage{pbalance}
\usepackage{enumitem}
\usepackage{booktabs}
\usepackage{array}
\usepackage{tabularx}
\usepackage{caption}
\usepackage{xcolor}
\usepackage{algorithm}
\usepackage[noend]{algpseudocode}

\newlist{rqlist}{itemize}{1}
\setlist[rqlist]{%
  label={},
  leftmargin=*,
  itemsep=1pt,
  topsep=2pt,
  parsep=0pt,
  partopsep=0pt
}

\providecommand{\floatnote}[1]{\par\vspace{2pt}\noindent\footnotesize #1}

\newcounter{ktstmt}

\newcommand{\ktlabel}[1]{\refstepcounter{ktstmt}\label{#1}}

\author{Saber Zerhoudi}
\orcid{0000-0003-2259-0462}
\affiliation{%
  \institution{University of Passau}
  \city{Passau}
  \country{Germany}
}
\email{szerhoudi@acm.org}

\author{Jelena Mitrovi\'{c}}
\orcid{0000-0003-3220-8749}
\affiliation{%
  \institution{University of Passau}
  \city{Passau}
  \country{Germany}
}
\email{jelena.mitrovic@uni-passau.de}

\author{Michael Granitzer}
\orcid{0000-0003-3566-5507}
\affiliation{%
  \institution{University of Passau}
  \city{Passau}
  \country{Germany}
}
\affiliation{%
  \institution{IT:U}
  \city{Linz}
  \country{Austria}
}
\email{michael.granitzer@uni-passau.de}

\begin{document}

\title{The Compaction Cliff in Long-Running AI Agent Memory}

\makeatletter
\setlength{\skip\footins}{9pt plus 2pt minus 1pt}

\newcommand{\acmrightssize}{\fontsize{8}{9.5}\selectfont}

\setlength{\emergencystretch}{1.5em} 

\settopmatter{printacmref=false}

\newcommand{\firstpagerights}[1]{%
  \begingroup
    \renewcommand\thefootnote{}%
    \footnotetext{%
      \acmrightssize
      \raggedright
      \setlength{\parskip}{0pt}%
      \setlength{\parindent}{0pt}%
      #1%
    }%
    \addtocounter{footnote}{0}%
  \endgroup
}
\makeatother

\begin{abstract}
A safety rule and an episodic log compete for the same tokens in an AI agent's context. When the budget overflows, both are summarized at the same rate; only the rule needs exact wording to remain enforceable. On 20 production agent configurations, Claude Code's \texttt{/compact} prompt on Sonnet 4.6 preserves 53\% of safety rules after one compaction round and 10\% after five. We name this the \emph{Compaction Cliff}. We address it with \emph{Knowledge Triage}, a framework that classifies each line of an agent's knowledge base by type and routes each type through its own retention policy. Three deterministic operators implement this triage across the three context-management operations: TypeCompact rewrites items in place under per-type fidelity, TypeDecompose partitions a topic too large to compact safely, replicating in-scope safety rules across partitions, and TypeRetrieve fetches items from external storage with in-scope rules pinned ahead of relevance. On five public corpora, TypeCompact preserves 2--4$\times$ more safety rules than the strongest single-shot LLM compactor at every ratio, with 96\% recall over five rounds. TypeDecompose reaches 0\% locality violations against 93\% under uniform partitioning. TypeRetrieve reaches 100\% recall@50 against 73\% for the best single-shot LLM retriever. On three downstream behavioral benchmarks, we outperform the production Sonnet compactor on medical compliance (paired McNemar $p < 10^{-8}$ on preservation, $N = 200$), the full-policy and hierarchical baselines on retail task pass rate ($p < 0.01$, $N = 115$), and the hierarchical compaction on the airline domain ($p = 0.024$). We release AgentArtifactCorpus (396{,}934 agent configurations from 54{,}628 public GitHub repositories), the classifier, and the reference implementation.
\end{abstract}



\begin{teaserfigure}
\centering
\includegraphics[width=\textwidth]{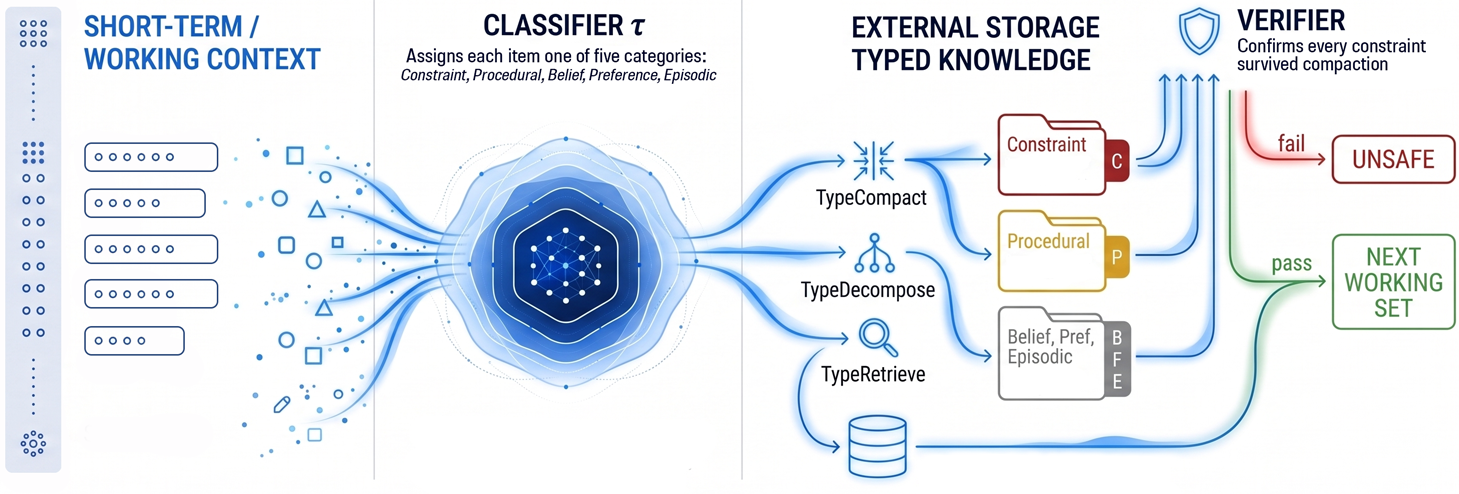}
\caption{Knowledge Triage. Classifier $\tau$ assigns each working-set item one of five types; three deterministic operators apply per-type policies: TypeCompact compresses without dropping safety rules, TypeDecompose splits large topics and duplicates rules that span them, TypeRetrieve returns applicable rules first. A verifier checks that every constraint survived; if any are missing, the output is flagged \textsc{unsafe}, otherwise it becomes the next working set.}
\label{fig:cycle}
\end{teaserfigure}

\maketitle
\enlargethispage{2\baselineskip}
\firstpagerights{%
  © ACM, 2026. This is the author's version of the work.\\
  The definitive version was published in:
  \emph{Proceedings of the 35th ACM International Conference on Information and Knowledge Management (CIKM '26), November 07--11, 2026, Rome, Italy}.\\
  DOI: \url{https://doi.org/10.1145/3799682.3840567}
}


\section{Introduction}
\label{sec:intro}

A medical AI agent reviews a patient's history and finds a critical line: \emph{``Patient is allergic to penicillin.''} Over a long session the context fills, the runtime triggers compaction, and the allergy line, buried among dozens of facts, is paraphrased or dropped. Three turns later the agent recommends amoxicillin, a penicillin-class antibiotic. Hierarchical truncation, the structural core of the summarize-and-retain compaction pattern deployed in production agents, preserves only 50\% of safety constraints on 50 real agent configurations (\S\ref{sec:exp:safety}); a 2026 community analysis of leaked Claude Code documents the same uniform-compaction pattern in production~\cite{claude_code_leak_2026}.

\looseness=-1 Agent runtimes apply three operations to keep the knowledge base inside its token budget: (i) \emph{Compaction} shrinks the working set in place by summarizing or pruning items, (ii) \emph{Decomposition} splits a topic too large to compact safely into sub-topics that each fit, (iii) \emph{Retrieval} pushes knowledge to external storage and pulls chunks back on demand.
The three share a single constraint, a finite budget over heterogeneous knowledge, yet are treated independently in the literature, with separate strategies for each. A second gap runs through all three: none of the production strategies condition retention on the \emph{type} of item. The closest typed-memory work~\cite{mars_2025} uses a single submodular utility across categories, so a safety constraint and a background belief are lost at the same rate under budget pressure.

Different kinds of knowledge tolerate different kinds of loss: a safety rule cannot be paraphrased without risking the qualifier that makes it actionable, a shell command can be rewritten only if it executes identically, a debugging trace can be collapsed to one sentence with no harm. One compression policy cannot serve all three, and \S\ref{sec:framework} proves this formally for each of the three operations.
We ask how many safety rules survive compaction in production, and find what we call the \textbf{Compaction Cliff}: under the production ``compress while keeping safety rules'' prompt on Sonnet 4.6, safety-rule recall holds at 53\% after one round and falls to 10\% by round five (\S\ref{sec:experiments}, Figure~\ref{tab:stability}). Prior work offers partial answers (information theory~\cite{liu_jsac_2024,he_iclr_2026}, belief-update calculi~\cite{agm_1985,park_2026}, typed memory~\cite{mars_2025}), but none treats compaction, decomposition, and retrieval as one problem with type-dependent safety guarantees. We address the cliff with \textbf{Knowledge Triage}, a framework that classifies each item once and routes each category through its own retention policy via three deterministic operators (TypeCompact, TypeDecompose, TypeRetrieve).
We contribute in five ways. (1) \emph{The Compaction Cliff}: a structural failure mode of type-blind compaction reproduced across four LLM compactor families and the strongest non-LLM compressor (\S\ref{sec:exp:safety}). (2) A five-type model of agent knowledge covering 97\% of real configuration content (\S\ref{sec:framework:model}, Table~\ref{tab:typology}). (3) Three operators with per-operator safety requirements that uniform strategies cannot satisfy, with proofs (\S\ref{sec:framework:operators}). (4) The Knowledge Triage framework: multi-fidelity retention lanes, a post-compaction guard verifier, and a SafetyMargin classifier that scores by counterfactual safety rather than grammatical form. (5) AgentArtifactCorpus, a new corpus of 396{,}934 agent knowledge artifacts from 54{,}628 public GitHub repositories under permissive licenses with 2{,}000 per-type annotations, together with end-to-end validation on five public corpora (\S\ref{sec:experiments}).

We release the dataset,\footnote{\url{https://huggingface.co/datasets/searchsim/AgentArtifactCorpus}} classifiers, and reference code.\footnote{\url{https://github.com/searchsim-org/cikm26-knowledge-triage}\label{fn:repo}}


\section{Related Work}
\label{sec:related}

Prior work treats compaction, decomposition, and retrieval as separate problems, evaluated separately; none formalize a per-type safety guarantee across all three. Table~\ref{tab:related-matrix} positions the closest agent-memory systems against the framework's properties.

\begin{table}[t]
\caption{Positioning of Knowledge Triage.}
\label{tab:related-matrix}
\small
\setlength{\tabcolsep}{4pt}
\begin{tabular*}{\columnwidth}{@{\extracolsep{\fill}}l c c c c c@{}}
\toprule
\textbf{System} & \textbf{Typed} & \textbf{Op C} & \textbf{Op D} & \textbf{Op R} & \textbf{Safety} \\
 & \textbf{model} & & & & \textbf{guarantee} \\
\midrule
MemGPT~\cite{memgpt_2023}             & --       & \checkmark & --        & \checkmark & --        \\
A-MEM~\cite{amem_2025}                & --       & --         & --        & \checkmark & --        \\
GraphRAG~\cite{graphrag_2024}         & --       & --         & --        & \checkmark & --        \\
LLMLingua-2~\cite{llmlingua2_2024}    & --       & \checkmark & --        & --        & --        \\
MaRS~\cite{mars_2025}                  & \checkmark & \checkmark & --      & --        & partial   \\
MemOS~\cite{memos_2025}                & partial  & \checkmark & --        & \checkmark & --        \\
\midrule
\textbf{Knowledge Triage}              & \textbf{\checkmark} & \textbf{\checkmark} & \textbf{\checkmark} & \textbf{\checkmark} & \textbf{per operator} \\
\bottomrule
\end{tabular*}
\floatnote{\emph{Typed model}: explicit per-item type assignment. \emph{Op C/D/R}: operator support for compaction, decomposition, retrieval. \emph{Safety guarantee}: formal per-operator preservation requirement that uniform strategies are shown to violate.}
\end{table}

\paragraph{Defining memories, from humans to agents.}
\looseness=-1 Cognitive science separated episodic from semantic memory~\cite{tulving_1972} and declarative from procedural knowledge~\cite{anderson_1983}. Agent systems carry these splits forward as engineering primitives: a paged hierarchy~\cite{memgpt_2023}, a Zettelkasten~\cite{amem_2025}, a reinforcement-learned consolidation~\cite{agemem_2026}, a memory-cube abstraction~\cite{memos_2025}, a tiered priority store~\cite{hmo_2026}, knowledge-graph triples with scope routing~\cite{hipporag_2024,graphrag_2024,park_2026}, and AGM-style belief revision~\cite{agm_1985,park_2026,falappa_2011}. MaRS~\cite{mars_2025} is the only prior system with both a typed model and structural safety properties, but it optimizes a single submodular utility across types, so a safety constraint can be lost at the same rate as a background belief under budget pressure. We lift these splits into the context-management layer: every working-set item carries one of five types, and each type carries its own retention rule across all three operations, giving constraint preservation under any feasible budget. Knowledge-graph memory is complementary: a graph can host items as nodes labeled by our classifier, with no extraction step because rules already live as natural-language strings in \texttt{AGENTS.md}, \texttt{CLAUDE.md}, and system prompts.

\paragraph{Why compaction loses safety rules.}
Two empirical lines explain the failure. Attention to long inputs is uneven: positional bias leaves middle content under-attended~\cite{lostmiddle_2024}, and LongBench~\cite{longbench_2024} / RULER~\cite{ruler_2024} document degradation across tasks. Abstractive summarization silently drops or inverts factual content~\cite{factcc_2020}. Hierarchical summarization inherits both losses. On top of model behavior, safety constraints are \emph{privileged instructions}~\cite{wallace_2024_hierarchy} that user-installed rules must dominate; Constitutional AI~\cite{constitutional_2022} writes such preferences as a fixed rule set; both are vulnerable to context manipulation~\cite{jailbroken_2023}, and R-Judge~\cite{rjudge_2024} benchmarks safety-risk awareness under that pressure. The compaction cliff is an \emph{unintentional} form of the same failure: type-blind hierarchical summarization deletes the rule the inference-time hierarchy was supposed to protect.

\paragraph{Improving compaction and context management.}
Production prompt compressors~\cite{llmlingua2_2024,recomp_2024,quitox_2025,xrag_2024} use a uniform fidelity target with no room for per-type guarantees; structuring and enriching the retrieved context is a related question~\cite{zerhoudi2026metadata}. KV-cache compression~\cite{streamingllm_2024,h2o_2023} acts at the inference-time cache, a different layer. Liu et al.~\cite{liu_jsac_2024} proved the rate-distortion theorem for composite sources under subsource-dependent fidelity; He et al.~\cite{he_iclr_2026} apply the lens to agentic design without separating fidelity by type. On retrieval, RAG~\cite{rag_2020}, Self-RAG~\cite{selfrag_2024}, and FLARE~\cite{flare_2023} score by relevance alone; agent-driven RAG such as PersonaRAG~\cite{zerhoudi2024personarag} adapts this to the user, and governed retrieval such as NuggetIndex~\cite{zerhoudi2026nuggetindex} filters records by validity before ranking, but neither conditions on constraint membership. We instantiate the composite-source theorem with knowledge types as subsources and exact-preservation distortion for constraints, so the guarantee survives partitioning and retrieval, complementary to inference-time hierarchy enforcement~\cite{wallace_2024_hierarchy}.

\begin{table*}[!t]
\caption{The five operational knowledge types of Knowledge Triage, with their distortion tolerance under context compression and illustrative examples drawn from several domains.}
\label{tab:typology}
\small
\setlength{\tabcolsep}{6pt}
\renewcommand{\arraystretch}{1.05}
\begin{tabular*}{\textwidth}{@{\extracolsep{\fill}}l p{0.22\textwidth} p{0.29\textwidth} p{0.31\textwidth}@{}}
\toprule
\textbf{Type} & \textbf{Definition} & \textbf{Distortion tolerance} & \textbf{Examples} \\
\midrule
Constraint (C) & Rule that bounds agent behavior; violation causes safety failure & Zero; any loss is unsafe & ``Never prescribe contraindicated drugs'' \\
Procedural (P) & Step-by-step instruction for a task & Semantic equivalence iff execution preserved & ``Run \texttt{pytest -x --tb=short} before merging'' \\
Belief (B) & Factual assertion treated as true & Bounded by semantic distance & ``Backend exposes a REST API over HTTPS'' \\
Preference (F) & Soft guideline & High; may be summarized or merged & ``Use 2-space indentation'' \\
Episodic (E) & Past event or observation & Free; may be discarded & ``Refactored auth module yesterday'' \\
\bottomrule
\end{tabular*}
\end{table*}


\section{Knowledge Triage Framework}
\label{sec:framework}

Like a hospital triaging mass-casualty patients red / yellow / green by tolerance to delay, an agent under a token budget triages its items by tolerance to distortion: a safety constraint must survive intact, a procedure may be rewritten when behavior is preserved, an episodic log may be dropped. Knowledge Triage operationalizes this principle through a typed knowledge model (\S\ref{sec:framework:model}), a classifier $\tau$ that assigns each item one of five types  (\S\ref{sec:framework:classifier}), and three operators with per-type retention policies (\S\ref{sec:framework:operators}): TypeCompact when items must fit a budget, TypeDecompose when no budget suffices, and TypeRetrieve for items outside the context. Each operator's safety requirement has a corresponding failure mode under type-blind strategies, proved in \S\ref{sec:framework:operators} and measured in \S\ref{sec:experiments}.

\subsection{Typed knowledge model}
\label{sec:framework:model}
We define five operational types derived from analysis of 396{,}934 agent knowledge artifacts scraped from source code repositories (\S\ref{sec:experiments} details the corpus). Table~\ref{tab:typology} summarizes the types and their compaction tolerance.

\looseness=-1 A flat list of typed items is enough for compaction, but decomposition and retrieval also need to know which items belong together. We extend the model with a topic hierarchy. A typed knowledge base\ktlabel{def:kb} is a tuple $K = (I, \tau, T, \pi, \sigma)$: a finite set of items $I$, the type assignment $\tau: I \to \{C, P, B, F, E\}$, a topic tree $T$, a leaf-mapping $\pi: I \to \mathrm{leaves}(T)$, and a scope function $\sigma: I_C \to 2^{\mathrm{leaves}(T)}$ that lists the leaf topics each constraint applies to (global-scope constraints have $\sigma(c) = \mathrm{leaves}(T)$). A working set $W \subseteq K$ is the subset loaded into context under a budget $|W| \leq B$. As a concrete illustration of $\sigma$, a coding-agent base may place \emph{Build}, \emph{Deployment}, and \emph{Database} under a project root, with the constraint \emph{``never modify production schema without a migration file''} sitting inside \emph{Database} but carrying $\sigma(c) = \mathrm{leaves}(T)$ because it applies wherever database operations occur.
\paragraph{Choice of granularity.} Five categories is empirically the coarsest partition that supports the per-operator safety arguments and the finest at which sub-divisions earn no new guarantees; \S\ref{sec:exp:coverage} reports the supporting coverage study on 2{,}000 annotated instructions.

\paragraph{Per-type distortion.} A single fidelity setting cannot serve all five types because they tolerate qualitatively different kinds of loss. Following Liu et al.~\cite{liu_jsac_2024}, each type carries its own distortion $d_t$: binary for constraints ($d_C = 0$ if $i' \models i$ and $\infty$ otherwise: the rule is intact or lost), behavior-preserving rewrites only for procedures, embedding distance $1 - \cos(e(i), e(i'))$ for beliefs and preferences (stricter for beliefs), and gist distance $1 - \mathrm{ROUGE}\text{-}\mathrm{L}(i, i')$ for episodic items, allowing summarization.

\subsection{Per-item type classification}
\label{sec:framework:classifier}
The classifier $\tau$ reads each working-set item and emits one of the five types from Table~\ref{tab:typology}. It is the only learned component in the framework, and every per-operator safety claim depends on it: a missed constraint sends the item to a soft retention lane where it can be paraphrased or dropped. We propose SafetyMargin as the default and three alternatives that trade safety recall for cost.

\paragraph{SafetyMargin (default).} An item is a constraint when removing it from the knowledge base would let some action become unsafe. SafetyMargin estimates this counterfactual directly: $\mathrm{margin}(i) = \max_{a \in \mathcal{A}} \big[P(\mathrm{unsafe} \mid a, K \setminus \{i\}) - P(\mathrm{unsafe} \mid a, K)\big]$ over a domain-conditioned action set. A single gpt-5.4-mini call\footnote{Full prompt text, regex pattern list, and cascade thresholds are in the released artifact.} scores this margin between 0 and 1; above 0.5 marks a Constraint. Counterfactual safety captures descriptively-stated rules (``the patient is allergic to X'') that grammatical-form classifiers miss, common in clinical, legal, and financial text; \S\ref{sec:exp:classifier} reports the per-phrasing comparison.

\paragraph{Alternatives.} A multi-stage \emph{selective cascade} (regex $\to$ encoder $\to$ gpt-5.4-mini $\to$ abstain-as-hard) routes regex-resolvable items through cheap stages first to reduce per-item LLM cost. A one-shot gpt-5.4-mini call under a fixed five-class prompt provides the simplest LLM baseline. Surface heuristics (regex, encoder, a distilled MiniLM~\cite{llmlingua2_2024}) cover settings where any LLM dependency is unacceptable. All four variants share the same five-class interface, run only at indexing time, and are never re-invoked by the operators in \S\ref{sec:framework:operators}, so the per-item cost is paid once and is constant in the number of compaction rounds, partitions written, and queries served; \S\ref{sec:exp:classifier} (Table~\ref{tab:classifier_ablation}) reports the ten-variant ablation.

\paragraph{Cascade construction.} The regex stage of the selective cascade matches each item against four families of hand-written patterns, authored against the gold-standard set of \S\ref{sec:exp:coverage} and tuned for precision so that a match resolves the item immediately. The families are tried in fixed precedence order, constraints first: constraint indicators (\emph{never}, \emph{must not}, capitalized markers such as \textsc{critical}, and \emph{always}/\emph{ensure}/\emph{require} combined with a modal verb); procedural indicators (commands opening with verbs such as \emph{run}, code blocks, and package-manager invocations); temporal markers for episodic items (\emph{yesterday}, \emph{recently}); and preference indicators (\emph{prefer}, \emph{ideally}). Unmatched items fall through to the encoder's prototype-centroid cosine (ten labeled examples per type); items scored below 0.55 confidence fall through to gpt-5.4-mini, and items still below 0.40 are routed to the hard lane as constraints. Notably, declarative safety statements carry none of these indicators. As a result, the cascade trails SafetyMargin on declarative phrasing.

\paragraph{Authority weighting.} We weight $\tau$'s per-item confidence by source authority $a: I \to \{\textsc{sys}, \textsc{dev}, \textsc{user}, \textsc{tool}, \textsc{ret}\}$ with $w_{\textsc{sys}}{=}1.5$, $w_{\textsc{dev}}{=}1.25$, $w_{\textsc{user}}{=}1.0$, $w_{\textsc{tool}}{=}0.75$, $w_{\textsc{ret}}{=}0.6$. The constraint risk is $\mathrm{risk}(i) = p(\tau(i){=}C) \cdot w_{a(i)}$, so a high-confidence system rule outweighs a low-confidence retrieved fragment.

\subsection{Per-operator context management}
\label{sec:framework:operators}

The three operators share a common shape: each takes a typed knowledge base, carries one safety requirement that depends on $\tau$, and has a known failure mode against $\tau$-blind strategies.

\subsubsection{Compaction.} A compaction operator $C: K \times \mathbb{N} \to K'$ maps a base and a budget to a compacted base $K'$ with $|K'| \leq B$. The safety requirement\ktlabel{inv:comp} is that every constraint in $K$ has a copy in $C(K)$ at zero distortion: $\forall i \in I_C, \exists i' \in C(K) : d_C(i, i') = 0$. This is feasible only above the minimum safe budget $B_{\min} = \sum_{i: \tau(i) \in \{C, P\}} |i|$; below it the system must expand the budget, relax safety, or fall through to decomposition. The three strategies deployed in production (hierarchical summarization, temporal windowing, and aggressive pruning, instantiated as the structural baselines of \S\ref{sec:exp:safety}) treat all items the same way and violate the requirement\ktlabel{thm:uniform-comp} as soon as a constraint falls inside the summarization window or below the relevance threshold. TypeCompact (Algorithm~\ref{alg:typecompact}) routes items into three fidelity lanes following Adaptive Focus Memory~\cite{adaptivefocus_2025}: constraints and procedures at \textsc{full}, beliefs and preferences at \textsc{compressed}, episodic items at \textsc{placeholder}. A deterministic verifier then extracts a canonical form (negation plus object phrase) from every constraint in the hard lane, checks it appears in the output, and restores the original or escalates to \textsc{Unsafe}~\cite{memmgmt_empirical_2025}.

\begin{algorithm}[t]
\caption{TypeCompact}
\label{alg:typecompact}
\begin{algorithmic}[1]
\Require $K$, budget $B$, classifier $h$, verifier $v$, authority $a$, thresholds $\theta_C, \theta_P$
\Ensure compacted $K'$ with $|K'| \leq B$, or \textsc{Unsafe}
\For{$i \in K$}\ \ $p_i \gets h(i,\,\mathrm{ctx}(i))$
  \If{$p_i(C)\,w_{a(i)} \geq \theta_C \lor \mathrm{abstain}(p_i)$}\ \ $H_C \gets H_C \cup \{i\}$;\ $g_i \gets \mathrm{guard}(i)$
  \ElsIf{$p_i(P) \geq \theta_P$}\ \ $H_P \gets H_P \cup \{i\}$
  \Else\ \ route $i$ to soft lane by argmax type
  \EndIf
\EndFor
\State $H \gets \mathrm{dedup}(H_C \cup H_P)$;\ \ \textbf{if} $\ell(H) > B$ \textbf{return} \textsc{Unsafe}
\State allocate $B - \ell(H)$ across soft lane: beliefs/prefs $\to$ \textsc{compressed}, episodic $\to$ \textsc{placeholder}
\State $K' \gets H \cup \mathrm{soft}$
\If{$v(\{g_i\}_{i \in H_C},\, K') = \textsc{fail}$}\ \ restore failed guards if budget permits, else \textbf{return} \textsc{Unsafe}\EndIf
\State \Return $K'$
\end{algorithmic}
\textit{\footnotesize Complexity: $O(|K|)$ classifier calls, $O(|H|\log|H|)$ allocation, $O(|I_C|)$ verification.}
\end{algorithm}

\subsubsection{Decomposition.} A decomposition operator $D: K \times \mathbb{N} \to \{K_1, \dots, K_m\}$ partitions $K$ into $m$ sub-bases such that $\bigcup K_i = K$ and each $K_i$ fits the per-partition budget. The safety requirement\ktlabel{inv:dec} is constraint locality: $\forall c \in I_C,\ \forall j \in [m]: \big(\exists\, i \in K_j : \pi(i) \in \sigma(c)\big) \Rightarrow c \in K_j$; every partition that holds an item covered by a constraint's scope must also hold the constraint. Heuristics that partition by topic similarity, item frequency, or token count ignore scope and violate locality\ktlabel{thm:uniform-dec} whenever a scoped constraint and a scoped item land in different partitions. TypeDecompose (Algorithm~\ref{alg:typedecompose}) replicates each constraint to every partition that intersects its scope, paying overhead proportional to the number of partitions the scope spans; for local constraints it is zero (empirical distribution in \S\ref{sec:experiments}).

\begin{algorithm}[t]
\caption{TypeDecompose}
\label{alg:typedecompose}
\begin{algorithmic}[1]
\Require knowledge base $K$, per-partition budget $B$
\Ensure partitions $\{K_1, \dots, K_m\}$, each $\leq B$
\State group $K$ by topic; chunk each group into sub-bases of size $\leq B$
\For{each constraint $c \in I_C$}
  \For{each partition $K_i$}
    \If{$\exists\, i' \in K_i : \pi(i') \in \sigma(c)$ and $c \notin K_i$}
      \State $K_i \gets K_i \cup \{c\}$ \Comment{replicate}
    \EndIf
  \EndFor
\EndFor
\State \Return $\{K_1, \dots, K_m\}$
\end{algorithmic}
\textit{\footnotesize Complexity: $O(|K|)$ grouping, $O(|I_C|\,m)$ replication, bounded by global-scope density.}
\end{algorithm}

\subsubsection{Retrieval.} Production retrievers (BM25, dense similarity, learned rerankers) rank by query relevance, which approximates necessity for non-constraint items but is unsafe for constraints; even scope-restricted retrieval~\cite{zerhoudi2026owlerlite}, which filters to a user-chosen collection, still ranks within that scope by relevance. The safety requirement\ktlabel{inv:ret} is priority: $\forall c \in I_C : \mathrm{inscope}(q, c) \Rightarrow c \in R_q$ regardless of similarity, where $\mathrm{inscope}(q, c) \Leftrightarrow \mathrm{topics}(q) \cap \sigma(c) \neq \emptyset$ tests whether the query's tagged topics intersect the constraint's scope; every in-scope constraint must appear in the result set even when its similarity score would not place it there. Any scoring function based on similarity alone can rank an in-scope constraint below the cutoff and violate priority\ktlabel{thm:uniform-ret}. TypeRetrieve (Algorithm~\ref{alg:typeretrieve}) pins in-scope constraints before allocating the residual budget by relevance.

\begin{algorithm}[t]
\caption{TypeRetrieve}
\label{alg:typeretrieve}
\begin{algorithmic}[1]
\Require corpus $K$, query $q$, budget $k$, scorer $r$, scope test $\mathrm{inscope}$
\State $C_q \gets \{c \in I_C : \mathrm{inscope}(q, c)\}$
\State $R_{\mathrm{others}} \gets \mathrm{topk}(\{i \in K \setminus C_q\},\, k - |C_q|,\, \mathrm{by} = r(q, \cdot))$
\State \Return $C_q \cup R_{\mathrm{others}}$
\end{algorithmic}
\textit{\footnotesize Complexity: $O(|I_C|)$ in-scope test plus scorer $r$; pinning adds no asymptotic overhead.}
\end{algorithm}

\paragraph{Composition.} The three operators compose into a single context-management cycle (Figure~\ref{fig:cycle}): the system attempts compaction first, and below $B_{\min}$, where compaction would drop a constraint, falls through to decomposition, pushing part of the base to external storage for TypeRetrieve to pull back on demand.\label{sec:framework:cycle}


\section{Empirical Validation}
\label{sec:experiments}

\looseness=-1 This section tests Knowledge Triage's three structural claims empirically and measures the deployment cost of the classifier they depend on. The claims are that the five-type model covers real agent knowledge (\S\ref{sec:framework:model}), that per-type retention preserves safety across the three operators where type-blind retention provably fails (\S\ref{sec:framework:operators}), and that artifact-level preservation translates to agent behavior on realistic tasks. Four research questions follow. \emph{RQ1:} how is real agent knowledge distributed across types, and what fraction is safety-critical (\S\ref{sec:exp:coverage})? \emph{RQ2:} does Knowledge Triage preserve safety-critical items where type-blind methods drop them, across all three operators (\S\ref{sec:exp:safety})? \emph{RQ3:} what does per-type retention cost in compute, classifier accuracy, and decomposition overhead (\S\ref{sec:exp:classifier})? \emph{RQ4:} does per-type retention change agent behavior on safety-critical benchmarks (\S\ref{sec:exp:behavior})?

\subsection{Setup}
\label{sec:exp:setup}

The six corpora in Table~\ref{tab:corpora} fill three roles. AgentArtifactCorpus (AAC), which we constructed for this paper, is the primary corpus for the typology and operator-level experiments; LongMemEval and BEIR scifact provide cross-corpus checks for the taxonomy and retrieval components; $\tau$-bench retail, $\tau$-bench airline, and SafetyMed provide end-to-end behavioral evaluation.

\begin{table}[t]
\caption{Corpora used in the empirical validation.}
\label{tab:corpora}
\small
\setlength{\tabcolsep}{4pt}
\renewcommand{\arraystretch}{1.05}
\begin{tabular*}{\columnwidth}{@{\extracolsep{\fill}}l l p{0.44\columnwidth}@{}}
\toprule
\textbf{Corpus} & \textbf{Scale} & \textbf{Role in this paper} \\
\midrule
\textbf{AAC (ours)}                                & 396{,}934 items   & Taxonomy, compaction, scope analysis (\S\S\ref{sec:exp:coverage}--\ref{sec:exp:safety}) \\
LongMemEval~\cite{longmemeval_2025}       & 500 turns         & Cross-corpus taxonomy validation (\S\ref{sec:exp:coverage}) \\
$\tau$-bench retail~\cite{taubench_2024}  & 115 tasks         & Retail-task behavioral rollouts (\S\ref{sec:exp:behavior}) \\
$\tau$-bench airline~\cite{taubench_2024} & 50 tasks          & Customer-service generalization test (\S\ref{sec:exp:behavior}) \\
BEIR scifact~\cite{beir_2021}             & 1{,}000 docs      & Retrieval baseline corpus (\S\ref{sec:exp:safety}) \\
\textbf{SafetyMed (ours)}                          & 200 scenarios     & Medical-compliance behavioral evaluation (\S\ref{sec:exp:behavior}) \\
\bottomrule
\end{tabular*}
\end{table}

AgentArtifactCorpus comprises 396{,}934 knowledge artifacts from 54{,}628 GitHub repositories spanning eight platforms (Claude, Cursor, Copilot, Windsurf, Continue, Aider, Codeium, Universal). Median artifact size is 1{,}201 bytes; primary languages are TypeScript (31.5\%), Python (18.7\%), and Go (18.2\%). Collection used the GitHub Code Search API with adaptive file-size partitioning to overcome the 1{,}000-result limit. Classification along four dimensions (platform, function, authorship, specificity) used regex heuristics with no LLM analysis to avoid downstream contamination.

\subsection{RQ1: Compaction Cliff Structure}
\label{sec:exp:coverage}

\paragraph{Motivation.} The Compaction Cliff in \S\ref{sec:exp:safety} was observed on one corpus under one production prompt. Whether the cliff is structural to type-blind retention or specific to that sample depends on two claims the framework relies on: that the five-type model in \S\ref{sec:framework:model} covers the items a deployed agent has to manage, and that the declarative-form failure mode treated in \S\ref{sec:exp:classifier} is widespread in real safety text. We test both.

\paragraph{Coverage.} We assembled a 2{,}000-item gold-standard set from 200 randomly sampled AAC repositories stratified by stars and language; two annotators labeled independently following the rubric in the released artifact, with second-pass adjudication against the same rubric, and 500 LongMemEval turns were labeled for cross-domain validation. We trained an NLP classifier combining structural features (mood, modal verbs, code-block presence, temporal markers), semantic features (sentence embeddings clustered with HDBSCAN), and keyword features (constraint indicators, procedural indicators, episodic markers), and evaluated it against the human labels. The typology covers 97\% of AAC instructions (Table~\ref{tab:typoresults}); the remaining 3\% are meta-instructions that reference other artifacts and carry no safety force. Constraint detection is the most stable type under domain shift, with F1 holding at 0.88 on both corpora; macro-F1 drops 3 points on LongMemEval (0.84 against 0.87 on AAC), with the drop concentrated in Preference and Belief.

\paragraph{Phrasing.} \looseness=-1 Coverage alone is not enough: if safety text in deployment were overwhelmingly imperative, the declarative failure mode would be a corner case. We tested how widespread declarative phrasing actually is by classifying 564 openFDA safety sentences (from \textsc{BOXED WARNING} and \textsc{CONTRAINDICATIONS} fields across 30 drugs) and 36 LegalBench contract-NLI explicit-identification clauses~\cite{legalbench_2023} into the four grammatical forms using a gpt-5.4-mini call. Declarative phrasing accounts for 49.8\% of openFDA safety text and 61.1\% of LegalBench safety clauses, against 22.5\% / 22.2\% imperative; conditional and passive forms make up the remainder. A 100-sentence stratified subsample re-labeled with a second independent judge gives Cohen's $\kappa = 0.88$ on the four-class label and $\kappa = 0.92$ on the binary declarative-vs-other (96\% observed agreement). Disagreement concentrates on the imperative-vs-modal-passive boundary while the declarative class itself is stable under the second judge, so the conclusion that declarative phrasing dominates is robust to inter-judge noise.

\paragraph{Discussion.} Both structural claims hold. The five-type model covers 97\% of agent items, so type-blind retention has no large uncovered class where the cliff could hide; declarative phrasing dominates safety-critical text in two independent deployment domains, so the grammatical-form failure mode is the deployment-relevant case. The Compaction Cliff therefore follows from the typology and phrasing distribution alone, and the \S\ref{sec:exp:safety} sample stands as a representative instance.

\begin{table}[t]
\caption{Knowledge type distribution and classifier performance on AgentArtifactCorpus ($n = 2{,}000$ annotations).}
\label{tab:typoresults}
\small
\begin{tabular*}{\columnwidth}{@{\extracolsep{\fill}}l c c c c@{}}
\toprule
\textbf{Type} & \textbf{\% of items} & \textbf{Precision} & \textbf{Recall} & \textbf{F1} \\
\midrule
Constraint  & 12.3 & 0.91 & 0.87 & 0.89 \\
Procedural  & 28.7 & 0.94 & 0.92 & 0.93 \\
Belief      & 31.4 & 0.86 & 0.84 & 0.85 \\
Preference  & 14.2 & 0.83 & 0.80 & 0.81 \\
Episodic    & 13.4 & 0.90 & 0.88 & 0.89 \\
\bottomrule
\end{tabular*}
\end{table}

\subsection{RQ2: Safety Preservation}
\label{sec:exp:safety}

\paragraph{Motivation.} Each operator in \S\ref{sec:framework:operators} carries a distinct safety requirement that type-blind strategies provably violate: constraint preservation under compaction, locality under decomposition, and priority under retrieval. For each requirement we ask two questions: do the strongest type-blind baselines satisfy it, and does the typed operator close the gap when each of its internal components is in place? We answer the first by running each operator against three or more baselines that cover the production landscape; we answer the second through component-removal ablations.

\paragraph{Compaction.} \looseness=-1 We tested ten strategies on 50 AAC configurations stratified by language, star count, and artifact count (622 constraints, 2{,}257 procedures). The strategies are: three structural baselines (\emph{hierarchical truncation}: keep the first $N$ tokens; \emph{temporal windowing}: keep the last $N$; \emph{aggressive pruning}: drop the items with the highest unigram-frequency tokens); LLMLingua-2~\cite{llmlingua2_2024}; four LLM compactors invoked with the production prompt ``compress to $N$ tokens, keep every safety rule and procedural command verbatim'' (gpt-5.4-nano, gpt-5.4-mini, Sonnet 4.6 behind Claude Code's \texttt{/compact}, Opus 4.7); MaRS-FL (our reimplementation of the MaRS~\cite{mars_2025} design choice this paper opposes; \S\ref{sec:discussion}); and TypeCompact. Each ran at three compression targets (50\%, 25\%, 10\%) on a 20-configuration subset, and was iterated for five rounds at 50\% per round to reflect a long-running agent. Constraint recall is the fraction of a configuration's classifier-labeled constraints that survive, measured by the key-token test of \S\ref{sec:exp:behavior} (validated against two human annotators, $\kappa = 0.92$ on Preserved/Lost). All strategies share the per-configuration budget; TypeCompact's pinned items count like any other.

\begin{figure}[t]
\centering
\includegraphics[width=.95\columnwidth]{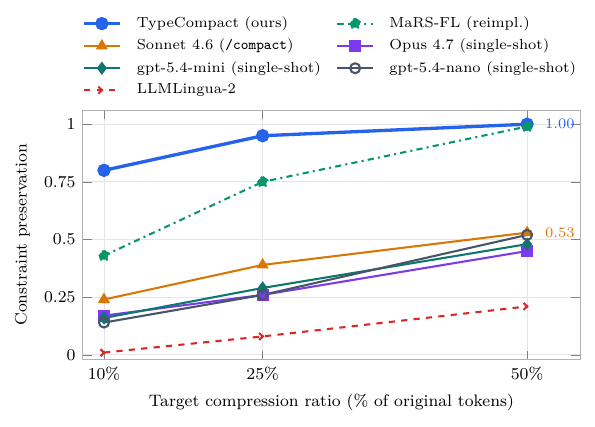}
\caption{Constraint preservation vs.\ target compression ratio on 20 AAC configurations. Structural baselines are tabulated at 50\% in Table~\ref{tab:comp}; full numerical detail in Table~\ref{tab:curves}.}
\label{fig:summary}
\end{figure}

\begin{figure}[t]
\centering
\includegraphics[width=0.8\columnwidth]{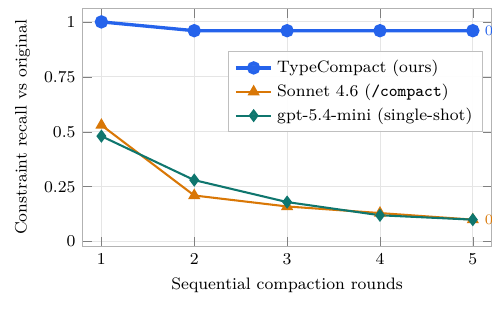}
\caption{Constraint recall after $N$ sequential compaction rounds at 50\% per round, 20 AAC configurations.}
\label{tab:stability}
\end{figure}

Across all eight type-blind strategies, the best single-round recall was 0.53 at 50\%, 0.39 at 25\%, 0.24 at 10\% (Figure~\ref{fig:summary}, Table~\ref{tab:curves}); five rounds of \texttt{/compact} drove recall from 0.53 to 0.10 (Figure~\ref{tab:stability}). The decay appeared for every LLM family and every structural baseline, so the loss is not specific to one prompt or one model. A type-blind compactor has no signal for which sentences are safety rules and summarizes them at the same rate as the surrounding text.

TypeCompact returned 1.00 / 0.95 / 0.80 constraint recall at 50 / 25 / 10\% on the same 50 configurations, stabilizing at 0.96 from the second round onward. This holds by construction: it pins every classifier-labeled constraint and procedure unchanged, so an item is lost only if the classifier mislabels it (bounded by recall, \S\ref{sec:exp:classifier}) or the joint footprint exceeds the budget (ablations below). Table~\ref{tab:comp} breaks the 50\% run down by type: TypeCompact trades belief and preference fidelity (0.50 / 0.51) for full constraint and procedural retention, while type-blind strategies spread it uniformly.

\paragraph{Compaction ablations.} TypeCompact has three components the type-blind baselines lack: indexing-time labels, a deterministic post-compaction verifier, and an \textsc{Unsafe} escalation when the budget cannot fit all pinned items. \emph{(i) Indexing-time labels.} LLMLingua-2 alone reached 0.55 / 0.18 / 0.02; adding a regex stage that re-inserts detected constraints after compression recovered to 0.83 / 0.65 / 0.60. This closes about half the gap; the other half closes only when the label is assigned before compression, because declarative-form rules carry no imperative marker for a regex to match. \emph{(ii) Verifier.} Across the run it recorded 27 restoration events (mean 0.46 per call, max 1) and never escalated to \textsc{Unsafe} at 50\%; without it the same runs would silently truncate the hard lane. \emph{(iii) Budget-boundary behavior.} On the 1.4\% of configurations whose constraint-plus-procedural density exceeded 40\%, $B_{\min}$ approached the 50\% budget; TypeCompact escalated to \textsc{Unsafe} on 28\% and reached 1.00 on the remaining 72\%, while the same configurations without the verifier reported apparent 1.00 recall but silently dropped a mean \textbf{57\%} of the constraints that should have been kept. Removing any single component puts a deployed agent below the 0.50 mark on this sample, so the result depends on all three together.

\begin{table}[t]
\caption{Preservation rate by knowledge type at 50\% compaction across 50 AAC configurations.}
\label{tab:comp}
\small
\setlength{\tabcolsep}{3pt}
\begin{tabular*}{\columnwidth}{@{\extracolsep{\fill}}l c c c c c c@{}}
\toprule
\textbf{Strategy} & \textbf{C} & \textbf{P} & \textbf{B} & \textbf{F} & \textbf{E} & \textbf{All} \\
\midrule
Hierarchical truncation       & 0.50 & 0.82 & \textbf{0.75} & 0.63 & 0.77 & 0.69 \\
Temporal windowing            & 0.58 & 0.80 & 0.64 & \textbf{0.78} & \textbf{0.91} & 0.74 \\
Aggressive pruning            & 0.51 & 0.75 & 0.67 & 0.66 & 0.84 & 0.69 \\
\textbf{TypeCompact}          & \textbf{1.00} & \textbf{1.00} & 0.50 & 0.51 & 0.84 & \textbf{0.77} \\
\bottomrule
\end{tabular*}
\floatnote{Items classified by $\tau$; higher is better. \emph{C}/\emph{P}/\emph{B}/\emph{F}/\emph{E} as in Table~\ref{tab:typology}; \emph{All}: unweighted mean.}
\end{table}

\begin{table}[t]
\caption{TypeCompact vs SOTA prompt compactors at three compression ratios, mean across 20 AAC configurations.}
\label{tab:curves}
\small
\begin{tabular*}{\columnwidth}{@{\extracolsep{\fill}}l c c c r r@{}}
\toprule
\textbf{Strategy} & \textbf{C} & \textbf{P} & \textbf{mF1} & \textbf{time} & \textbf{tok} \\
\midrule
\multicolumn{6}{l}{\textit{50\% compression}} \\
\textbf{TypeCompact}        & \textbf{1.00} & \textbf{0.92} & \textbf{0.75} & \textbf{0.0s} & \textbf{0} \\
LLMLingua-2                 & 0.21 & 0.72 & 0.63 & 0.5s & 0 \\
MaRS-FL (ours, no public impl)  & 0.99 & 0.89 & 0.69 & 0.5s & 0 \\
gpt-5.4-nano (single-shot)      & 0.52 & 0.80 & 0.69 & 9.9s  & 56K \\
gpt-5.4-mini (single-shot)      & 0.48 & 0.77 & 0.68 & 7.3s  & 57K \\
Sonnet 4.6 (\texttt{/compact}) & 0.53 & 0.83 & 0.73 & 73.9s & 48K \\
Opus 4.7 (single-shot)          & 0.45 & 0.81 & 0.72 & 31.1s & 50K \\
\midrule
\multicolumn{6}{l}{\textit{25\% compression}} \\
\textbf{TypeCompact}        & \textbf{0.95} & \textbf{0.71} & \textbf{0.57} & \textbf{0.0s} & \textbf{0} \\
LLMLingua-2                 & 0.08 & 0.61 & 0.46 & 0.5s & 0 \\
MaRS-FL (ours, no public impl)  & 0.75 & 0.61 & 0.51 & 0.2s & 0 \\
gpt-5.4-nano (single-shot)      & 0.26 & 0.60 & 0.50 & 9.3s  & 53K \\
gpt-5.4-mini (single-shot)      & 0.29 & 0.62 & 0.49 & 7.5s  & 56K \\
Sonnet 4.6 (\texttt{/compact}) & 0.39 & 0.65 & 0.51 & 166.2s & 42K \\
Opus 4.7 (single-shot)          & 0.26 & 0.68 & 0.51 & 35.6s & 43K \\
\midrule
\multicolumn{6}{l}{\textit{10\% compression}} \\
\textbf{TypeCompact}        & \textbf{0.80} & 0.47 & \textbf{0.38} & \textbf{0.0s} & \textbf{0} \\
LLMLingua-2                 & 0.01 & 0.27 & 0.14 & 0.5s & 0 \\
MaRS-FL (ours, no public impl)  & 0.43 & 0.22 & 0.27 & 0.02s & 0 \\
gpt-5.4-nano (single-shot)      & 0.14 & 0.43 & 0.33 & 5.6s  & 43K \\
gpt-5.4-mini (single-shot)      & 0.16 & 0.47 & 0.32 & 5.1s  & 46K \\
Sonnet 4.6 (\texttt{/compact}) & 0.24 & \textbf{0.54} & 0.30 & 420.3s & 36K \\
Opus 4.7 (single-shot)          & 0.17 & 0.52 & 0.31 & 22.8s & 37K \\
\bottomrule
\end{tabular*}
\floatnote{Outputs truncated to budget tokens before scoring; Sonnet 4.6 row uses Claude Code's \texttt{/compact}; \emph{MaRS-FL} reimplements MaRS~\cite{mars_2025} (\S\ref{sec:discussion}). \emph{C}: constraint preservation, \emph{P}: procedural, \emph{mF1}: macro F1, \emph{time}/\emph{tok}: mean latency / LLM tokens per config.}
\end{table}

\label{sec:llmlingua}

\paragraph{Decomposition.} The second operator requirement, locality, is tested on 200 AAC configurations whose section headers serve as topic markers. A constraint is global-scope if it appears in a section labeled \emph{Rules}, \emph{Constraints}, \emph{Safety}, or \emph{Project Conventions} (the conventional naming across the sample), so the topic-aligned baseline and TypeDecompose consume the same scope signal. The three baselines are \texttt{chunk\_by\_tokens} (sequential by token count), \texttt{chunk\_by\_topic} (by section header), and \texttt{chunk\_by\_paragraph\_budget} (greedy paragraph packing), each at a per-partition budget of 25\% of total tokens. A locality violation is a partition that holds a scoped item without the constraint that scopes it; we report the mean violation rate and the fraction of configurations with at least one violation. The strongest baseline (topic-aligned) reaches 13\% mean violation but still produces at least one violation in 40\% of configurations; the other two baselines reach 32\% / 93\% (Table~\ref{tab:dec}). TypeDecompose reaches zero violations at 14.5\% mean token overhead.

Extremes drive the 14.5\% mean: the median is 0\% (74\% of configurations have no global-scope constraints), the 90th percentile 21\%, the worst case 219\%; the shape persists across classifier variants with only the mean shifting with the flag rate, so the extremes come from the input distribution. A scaling test on a 3{,}000-item topic runs recursive TypeDecompose ($T = 200$ split threshold, KMeans over embeddings), producing 29 leaves of mean size 103; held-out retrieval at $k = 5$ reaches 94\% recall against the 98\% flat-scan bound while examining 92 of 3{,}000 items per query, a 33$\times$ reduction.

\begin{table}[t]
\caption{Decomposition constraint-locality on 200 AAC configurations, 25\%-of-total partition budget.}
\label{tab:dec}
\small
\setlength{\tabcolsep}{3pt}
\begin{tabular*}{\columnwidth}{@{\extracolsep{\fill}}l c c c@{}}
\toprule
\textbf{Strategy} & \textbf{Mean viol.} & \textbf{Configs viol.} & \textbf{Overhead} \\
\midrule
chunk\_by\_tokens             & 32\% & 93\% & 0\% \\
chunk\_by\_topic              & 13\% & 40\% & 0\% \\
chunk\_by\_paragraph          & 32\% & 93\% & 0\% \\
\textbf{TypeDecompose}        & \textbf{0\%}  & \textbf{0\%} & +14.5\% \\
\bottomrule
\end{tabular*}
\floatnote{Items reclassified by the selective classifier $\tau$ before partitioning. \emph{Configs viol.} fraction of configurations with a locality violation. \emph{Overhead} replication cost in tokens.}
\end{table}

\paragraph{Retrieval.} The third operator requirement, priority, asks whether in-scope constraints are returned ahead of merely-relevant items; score-only retrievers cannot enforce this because the score does not encode constraint membership. The corpus is 1{,}000 BEIR scifact documents as distractors plus 33 $\tau$-bench retail-policy chunks; the selective classifier $\tau$ flags 22 of the 33 as constraints (regex flags 3). The 50 queries are hand-written against the $\tau$-bench retail rubric; on average 12 of the 22 retail-policy constraints are in scope per query (in-scope = constraint's topic field matches at least one of the query's tagged topics). We compare three non-LLM retrievers (BM25; dense with \texttt{octen-embedding-8b} 4{,}096-dim cosine; cross-encoder rerank of dense top-50 with \texttt{qwen3-reranker-4b}), three single-shot LLM retrievers given the dense top-100 pool (gpt-5.4-nano, gpt-5.4-mini, Sonnet 4.6), and TypeRetrieve on the dense retriever (in-scope constraints pinned first, residual budget filled by relevance). Recall@$k$ for $k \in \{5, 10, 20, 50\}$ is the fraction of in-scope constraints in the top $k$. TypeRetrieve reaches 96\% recall@20 and 100\% recall@50 across every retriever (Table~\ref{tab:ret}); the strongest single-shot LLM retriever (Sonnet 4.6) reaches 61\% and 73\% at the same budgets. At $k = 5$ both approaches are close to the structural bound $\min(5 / 12, 1) \approx 42\%$; at $k = 50$ TypeRetrieve fills the in-scope set deterministically while score-only retrievers continue to spend the residual budget on high-similarity non-constraint items. TypeRetrieve uses \textbf{zero LLM tokens per query} against 5{,}776--6{,}741 for the single-shot LLMs, because pinning is a database lookup.

\begin{table*}[t]
\begin{minipage}[t]{0.58\textwidth}
\caption{In-scope constraint recall@$k$ on a 1{,}033-item mixed corpus, 50 queries.}
\label{tab:ret}
\centering
\small
\setlength{\tabcolsep}{3pt}
\begin{tabular*}{\linewidth}{@{\extracolsep{\fill}}l l c c c c r r@{}}
\toprule
\textbf{Retriever} & \textbf{Variant} & $@5$ & $@10$ & $@20$ & $@50$ & \textbf{sec/q} & \textbf{tok/q} \\
\midrule
\multicolumn{8}{@{}l}{\textit{Non-LLM retrievers + TypeRetrieve overlay}} \\
BM25            & baseline     & 32\%  & 38\%  & 43\%  & 49\% & 0.0 & 0 \\
Dense (octen)   & baseline     & 34\%  & 45\%  & 67\%  & 91\% & 0.0 & 0 \\
Reranked dense  & baseline     & 25\%  & 36\%  & 55\%  & 91\% & 0.0 & 0 \\
\textbf{TypeRetrieve}    & dense & \textbf{40\%} & \textbf{57\%} & \textbf{96\%} & \textbf{100\%} & \textbf{0.0} & \textbf{0} \\
\midrule
\multicolumn{8}{@{}l}{\textit{Single-shot LLM retrievers (full pipeline; given top-100 candidate pool)}} \\
gpt-5.4-nano (one-shot)   & baseline & 34\%  & 43\%  & 51\%  & 53\%  & 1.0 & 5{,}776 \\
gpt-5.4-mini (one-shot)   & baseline & 34\%  & 46\%  & 64\%  & 70\%  & 1.2 & 5{,}800 \\
Sonnet 4.6 (one-shot)     & baseline & 34\%  & 47\%  & 61\%  & 73\%  & 5.9 & 6{,}741 \\
\bottomrule
\end{tabular*}
\floatnote{Cascade $\tau$ flags 22 of 33 retail-policy chunks as constraints, 12 in scope per query on average. Single-shot LLM retrievers select the top-50 from the dense top-100 pool; that ranking is evaluated at all four budgets.}
\end{minipage}
\hfill
\begin{minipage}[t]{0.40\textwidth}
\captionof{figure}{Constraint recall by phrasing form, 50 base rules $\times$ 4 forms = 200 items.}
\label{fig:adv_form}
\centering
\includegraphics[width=\linewidth]{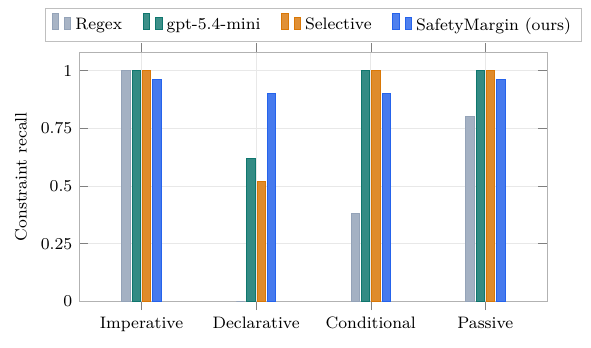}
\end{minipage}
\end{table*}

\paragraph{Discussion.} Why does typing close the gap on three operators? The type-blind alternatives share one blind spot: uniform compactors, topic-only partitioners, and score-only retrievers all lack a separate channel for constraint membership. LLMLingua-2 weights tokens by entropy, so rare imperative tokens such as \emph{never} are pruned alongside neutral fillers (\emph{``Patient is allergic to penicillin. Never prescribe\dots''} compresses at rate 0.5 with the negation dropped); single-shot LLM compactors interleave rules with other items, so truncation drops them at the same rate as the rest, and the multi-round decay (Figure~\ref{tab:stability}) is the same failure mode iterated; score-only retrievers spend the residual budget on high-similarity non-constraint items. The four ablations (regex re-insertion, verifier, budget-boundary behavior, replication-overhead distribution) each remove one component of the typed pipeline, and the gap re-opens in every case. Restoring that channel closes the gap for each operator: 100\% constraint preservation under feasible budgets, zero locality violations at 14.5\% mean overhead, and 100\% in-scope recall@50 across every retriever.

\subsection{RQ3: Classifier Cost}
\label{sec:exp:classifier}

\paragraph{Motivation.} All three operator claims in \S\ref{sec:framework:operators} depend on the classifier $\tau$, so the cost of per-type retention is the cost of running $\tau$ at scale paired with the recall it achieves at that operating point. We ask two questions: what does the cost-recall trade-off look like across cheap-to-expensive classifier variants on real agent text, and does that ranking hold when the safety text is grammatically declarative (the deployment-relevant case identified in \S\ref{sec:exp:coverage})? We close with a third measurement, authority weighting on mixed-source items, that informs TypeRetrieve.

\paragraph{Setup.} We evaluate ten classifier variants on a 200-item AAC test set. The variants are: regex, encoder (octen-embedding-8b prototype-centroid cosine), regex+encoder, distilled MiniLM~\cite{llmlingua2_2024}, four single-LLM one-shot variants (gpt-5.4-nano, gpt-5.4-mini, Sonnet 4.6, Opus 4.7), the multi-stage selective cascade (regex $\to$ encoder $\to$ gpt-5.4-mini), and selective + abstain-as-hard. SafetyMargin replaces grammatical classification with counterfactual safety-margin estimation: one gpt-5.4-mini call scores each item between 0 and 1, and scores above 0.5 mark it as a Constraint.

\paragraph{Reference labels.} The reference labels come from gpt-5.4-mini under the five-type definitions, so recall numbers for gpt-5.4-mini-based variants measure self-consistency and act as upper bounds. The adversarial-phrasing test below does not use these labels: its items rewrite author-written safety rules, so every label is known in advance. An independent annotator also re-labeled the 200-item set (\S\ref{sec:exp:behavior}). Five-class Cohen's $\kappa$ is 0.45, with disagreement concentrated in procedural-vs-belief decisions on code snippets; on the binary constraint-vs-other decision, $\kappa$ is 0.79 (95\% agreement). Against this human reference, gpt-5.4-mini reaches constraint precision 0.77, recall 0.88, and F1 0.82; against its own labels it reaches 0.84 (Table~\ref{tab:classifier_ablation}). For the safety claims, only the constraint-vs-other decision matters: a procedural-vs-belief confusion changes how an item is compressed and does not affect constraint retention.

\begin{table}[t]
\caption{Classifier variants on the 200-item AAC test set, reference labels from gpt-5.4-mini.}
\label{tab:classifier_ablation}
\small
\setlength{\tabcolsep}{3pt}
\begin{tabular*}{\columnwidth}{@{\extracolsep{\fill}}l c c c c r@{}}
\toprule
\textbf{Variant} & \textbf{C-Rec.} & \textbf{C-F1} & \textbf{mF1} & \textbf{Acc} & \textbf{Lat.} \\
\midrule
Regex only                          & 0.60 & 0.63 & 0.35 & 0.53 & $<$1\,ms \\
Encoder only                        & 0.27 & 0.32 & 0.33 & 0.41 & 387\,ms \\
Regex + Encoder                     & 0.70 & 0.63 & 0.40 & 0.46 & 266\,ms \\
Distilled MiniLM+LR~\cite{llmlingua2_2024}     & 0.77 & 0.61 & 0.44 & 0.52 & $\sim$5\,ms \\
\midrule
gpt-5.4-nano             & \textbf{0.93} & 0.74 & 0.62 & 0.69 & 684\,ms \\
\textbf{gpt-5.4-mini}    & 0.87 & \textbf{0.84} & 0.68 & \textbf{0.75} & 667\,ms \\
Sonnet 4.6               & 0.83 & 0.77 & 0.50 & 0.56 & 3{,}452\,ms \\
Opus 4.7                 & 0.73 & 0.80 & 0.61 & 0.67 & 3{,}853\,ms \\
Qwen3-Next-80B (OSS)     & 0.87 & 0.83 & 0.64 & 0.73 & 606\,ms \\
\midrule
Selective cascade           & \textbf{0.93} & 0.80 & 0.59 & \textbf{0.75} & 1{,}096\,ms \\
Selective + abstain         & 0.90 & 0.73 & \textbf{0.69} & 0.73 & 1{,}008\,ms \\
\bottomrule
\end{tabular*}
\floatnote{Single-LLM rows are one-shot calls. \emph{C-Rec.} constraint recall (safety-critical: false negatives are missed safety rules). \emph{mF1} macro F1. \emph{Lat.} per-call latency. Rows built on gpt-5.4-mini are self-consistency upper bounds.}
\end{table}

\paragraph{Cost-recall on AAC.} The cheapest variants (regex, encoder, hybrid, distilled MiniLM) reach 0.27--0.77 constraint recall (Table~\ref{tab:classifier_ablation}). The four single-shot LLMs reach between 0.73 and 0.93: gpt-5.4-nano matches the selective cascade at 0.93, gpt-5.4-mini reaches 0.87 against its own reference labels, and Opus 4.7 drops to 0.73 because its reasoning trace overflows the single-word output format. gpt-5.4-mini gives the best joint trade-off on AAC (constraint F1 0.84, macro F1 0.68, accuracy 0.75, 667ms per call). At 200 items the per-cell standard error is 3 to 5 points, so only large differences are reliable: the LLM variants beat the cheap heuristics and Opus 4.7 falls below the other LLMs, while gaps of a few points are not.

\paragraph{Adversarial phrasing.} \looseness=-1 The AAC ranking in Table~\ref{tab:classifier_ablation} is measured on mostly imperative text; we tested whether the same ranking holds when the safety text is declarative. Rewriting 50 author-written safety rules in four grammatical forms (imperative, declarative, conditional, passive) gives 200 items whose correct labels are known in advance (Figure~\ref{fig:adv_form}). The LLM rows handle imperative, conditional, and passive forms cleanly (1.00 recall each) but lose 38--48 absolute points on declarative phrasing: ``the patient is allergic to penicillin'' is recognized as a constraint by gpt-5.4-mini 62\% of the time, by the selective cascade 52\%, by regex 0\%. SafetyMargin lifts declarative recall from \textbf{0.62 to 0.90} (+28 points), at a cost of 4--10 points on the easier forms; mean recall across the four forms rises from 0.91 to 0.93, with 880ms latency per item against 667ms for the one-shot grammatical classifier.

\paragraph{Choosing a classifier.} The two tests give two recommendations. On AAC (imperative-heavy), the cost-recall frontier is \{Regex, Distilled MiniLM, gpt-5.4-mini, gpt-5.4-nano\}, and SafetyMargin sits below it because its declarative advantage is invisible on imperative text. On worst-case recall across phrasing forms, the frontier is \{Regex, gpt-5.4-mini, SafetyMargin\}, and SafetyMargin is recall-maximizing because grammatical classifiers collapse on declarative form (0.52--0.62) while SafetyMargin holds at 0.90. \S\ref{sec:exp:coverage} shows declarative phrasing dominates clinical, legal, and financial safety text, so deployments in those domains follow the second frontier.

\paragraph{Authority weighting on mixed sources.} The risk score of \S\ref{sec:framework:classifier}, evaluated against unweighted full-pin TypeRetrieve on the 1{,}033-item retrieval corpus, reaches recall@\{5, 10, 20, 50\} of 0.40 / 0.58 / 0.93 / 1.00 (full-pin 0.40 / 0.57 / 0.96 / 1.00; dense 0.34 / 0.44 / 0.69 / 0.93). The weighting keeps recall@50 at 1.00 with a smaller pinned footprint, useful under a tight per-partition budget.

\paragraph{Discussion.} The cost-recall gap is large but not monotone in model size: gpt-5.4-nano matches the selective cascade, while the larger Opus 4.7 underperforms it because of the output-format issue. The AAC ranking does not hold under declarative phrasing, where counterfactual classification (SafetyMargin) lifts recall by 28 absolute points. We therefore recommend SafetyMargin as the default for mixed-phrasing deployments, the selective cascade as the cost-saving alternative for imperative-heavy text, the one-shot LLM as the simplest baseline, and regex for no-LLM deployments.

\subsection{RQ4: Downstream Behavior}
\label{sec:exp:behavior}

\paragraph{Motivation.} The first three RQs measure safety preservation at the artifact level under an automated key-token metric. Two further questions remain before per-type retention is deployable: does the automated metric agree with human judgment, and does artifact-level preservation translate to safer agent behavior when a compacted policy is used in a live task? We answer them in turn.

\paragraph{Human verification.} Two annotators, using the released 4-class rubric, independently labeled a 100-cell sample from the \S\ref{sec:exp:safety} compaction experiment; 80 cells were double-labeled for agreement, 20 single-labeled, with $\kappa$ computed before any reconciliation. Cohen's $\kappa$ is 0.77 on the 4-class label (85\% agreement) and 0.92 on the binary preserved-vs-lost decision (96\%). Annotator consensus agrees with the automated decision on 79\% of cells; disagreements concentrate on the \emph{Weakened} class (28\% of automated-Preserved cells, mostly at 10\% compression where key tokens survive but a qualifier is dropped). Humans are therefore stricter, and this costs the type-blind compactors far more than TypeCompact: TypeCompact's preserved rate falls 5 points (91\% to 86\%) under human judgment, against 15 for temporal windowing (65\% to 50\%), 23 for hierarchical truncation (64\% to 41\%), and 30 for aggressive pruning (71\% to 41\%). The artifact-level gap thus widens under human judgment.

\paragraph{Behavioral benchmarks.} \looseness=-1 The three benchmarks cover the three policy conditions compaction faces in deployment: \emph{SafetyMed} for long policies dominated by declarative safety facts; \emph{$\tau$-bench retail} for short policies dominated by imperative rules; \emph{$\tau$-bench airline} for cross-domain generalization within customer service. The protocol is shared: each benchmark runs paired (full / type-blind compactor / TypeCompact) rollouts with a paired McNemar exact two-sided test.

\emph{SafetyMed} is a 200-scenario benchmark we constructed from MedQA~\cite{medqa_2020}, pairing each scenario with FDA \textsc{BOXED WARNING} and \textsc{CONTRAINDICATIONS} text via the openFDA API\footnote{Construction script and full benchmark in the released artifact; FDA labels are public-domain, MedQA uses its original license.}; each scenario embeds a gold REFUSE/PROCEED action the agent must produce. Across the four conditions (full / hierarchical / single-shot Sonnet 4.6 / TypeCompact + SafetyMargin), TypeCompact reaches 97.0\% pass and 95.5\% preservation, against 98.0\% / 93.5\% for hierarchical, 92.5\% / 81.0\% for Sonnet, and 96.5\% / 100.0\% for the full-policy ceiling. TypeCompact outperforms the production single-shot LLM compactor on pass rate ($p = 0.022$) and preservation ($p = 3.7 \times 10^{-9}$, a 14.5-point lead), and Sonnet's preservation drop concentrates on declaratively-written boxed-warning sentences, the grammatical-classifier failure mode \S\ref{sec:exp:classifier} traces to declarative phrasing. TypeCompact ties hierarchical statistically on pass and preservation and falls 4.5 preservation points below the full-policy ceiling ($p = 0.0039$): nine items the SafetyMargin classifier missed at its 0.93 operating recall. Per-type retention preserves no more constraints than the classifier detects.

\looseness=-1 \emph{$\tau$-bench retail}~\cite{taubench_2024} stresses the complementary case: a 1{,}338-token policy with little room to compress and mostly imperative rules. We ran 1{,}035 paired rollouts over 115 tasks $\times$ three model families (gpt-5.4-nano, gpt-5.4-mini, qwen3-next-80b) $\times$ three conditions (full, 50\%-head-truncated hierarchical at 669 tokens, and TypeCompact at 1{,}136 tokens after pinning 22 constraints and 6 procedures). Mean pass is 37.7\% TypeCompact, 28.6\% full, 29.2\% hierarchical (Table~\ref{tab:behav}); both leads cross significance under paired McNemar at $N = 329$ observations ($p = 0.0026$ vs full, $p = 0.0051$ vs hierarchical). The gain concentrates on gpt-5.4-nano, the smallest model, where the agent benefits most from pinning the relevant rules rather than relying on the model to locate them inside the full policy.

\emph{$\tau$-bench airline} tests whether the retail finding extends to a second customer-service domain. We ran the same protocol on the airline split (50 tasks, 6{,}155-character policy reduced to 718 tokens hierarchical and 647 tokens TypeCompact). Over the 117 (model, task) pairs that completed under all three conditions, mean pass is 26.5\% TypeCompact, 34.2\% full, 15.4\% hierarchical. TypeCompact outperforms hierarchical (paired McNemar $p = 0.024$, an 11.1-point lead) and ties the full-policy ceiling statistically ($p = 0.14$). The airline domain is harder than retail across the board, with lower pass rates everywhere and gpt-5.4-nano erroring on 23 of 50 full-policy cells; under that pressure, the compaction step loses some context the full policy gives the agent. The deployment-relevant comparison is against the type-blind compactor a production system would use (hierarchical truncation), and TypeCompact prevails on it.

\paragraph{Discussion.} The two motivation questions resolve in the same direction. The automated metric does not over-credit TypeCompact: under stricter human judgment the artifact-level gap to the type-blind methods widens, with TypeCompact losing 5 absolute points against 15--30 for the others. Across the three behavioral benchmarks, TypeCompact outperforms the deployable type-blind compactor in every case (Sonnet on SafetyMed, hierarchical on retail, hierarchical on airline), and the residual gap to the full-policy ceiling tracks classifier recall: small on retail (imperative rules, classifier reliable), 4.5 preservation points on SafetyMed (where the classifier missed nine declarative items at its 0.93 measured recall), within sampling noise on airline. The asymmetric retail-vs-airline outcome (TypeCompact outperforms full on retail, ties full on airline) matches what \S\ref{sec:framework:operators} proves about imperfect classification: the per-type retention property bounds its cost without eliminating it, so harder domains move TypeCompact closer to the full-policy ceiling without crossing below the deployable baseline. The retail comparison is equal-policy but not equal-tokens: TypeCompact's budget-adaptive footprint retains 1{,}136 tokens against 669 for hierarchical truncation, so part of the retail gain may reflect retained context rather than typing alone. Two controls bound this effect: the \S\ref{sec:exp:safety} experiments are budget-matched (pinned items count against the shared budget), and on airline the footprint reverses (647 against 718 tokens) while TypeCompact still leads by 11.1 points. A token-matched behavioral control is an open direction.

\begin{table}[t]
\caption{$\tau$-bench retail pass rate, $N = 115$ stratified tasks per cell, 1{,}035 paired rollouts (1{,}019 completed without error).}
\label{tab:behav}
\small
\setlength{\tabcolsep}{3pt}
\begin{tabular*}{\columnwidth}{@{\extracolsep{\fill}}l c c c c c@{}}
\toprule
\textbf{Model} & \textbf{Full} & \textbf{Hier.} & \textbf{TypeC.} & \textbf{$p$ vs full} & \textbf{$p$ vs hier.} \\
\midrule
gpt-5.4-nano        & 22.5\% & 21.6\% & \textbf{36.0\%} & \textbf{0.011} & \textbf{0.007} \\
gpt-5.4-mini        & 27.3\% & 29.1\% & \textbf{35.5\%} & 0.176 & 0.324 \\
qwen3-next-80b      & 36.1\% & 37.0\% & \textbf{41.7\%} & 0.345 & 0.424 \\
\midrule
\textbf{Paired overall} & 28.6\% & 29.2\% & \textbf{37.7\%} & \textbf{0.003} & \textbf{0.005} \\
\bottomrule
\end{tabular*}
\floatnote{P-values are exact two-sided paired McNemar. \emph{Full} 1{,}338-token policy; \emph{Hier.} 50\% hierarchical truncation (669 tokens); \emph{TypeC.} TypeCompact (1{,}136 tokens after pinning 22 constraints and 6 procedures). \emph{Paired overall} aggregates the 329 (model, task) pairs that completed all three conditions.}
\end{table}

\section{Discussion and Limitations}
\label{sec:discussion}

\paragraph{Implications.} The Compaction Cliff is structural to type-blind retention: it appears with the same shape on every LLM family and structural baseline we tested (\S\ref{sec:exp:safety}). Per-type retention removes it by conditioning on each item's subsource, which surface features alone do not reveal~\cite{liu_jsac_2024}; the same conditioning yields locality under decomposition and priority under retrieval, exposing a constraint-membership signal that score-only retrievers and topic-similarity partitioners do not observe. The framing is a typed-source rate-distortion account with three safety theorems (\S\ref{sec:framework:operators}) that link agent memory to information theory~\cite{liu_jsac_2024}, belief revision~\cite{agm_1985,park_2026}, and instruction-hierarchy safety~\cite{wallace_2024_hierarchy,jailbroken_2023} through a single mechanism.

\paragraph{Practical recommendations.} A production runtime can place a type classifier upstream of its compaction, retrieval, and decomposition modules and route flagged constraints through an exact-preservation path, at one classifier call per item paid once at indexing. A configuration author can read type composition as a quality signal (configurations dominated by preferences and recent history offer weaker safety than those with explicit constraints and procedures), and a multi-agent planner that summarizes before delegating inherits the cliff unless TypeCompact is applied at the summarization step.

\paragraph{Conditions on the guarantee.} Under a feasible budget the operators preserve every constraint the classifier flags. A constraint is lost only when the classifier misses it, so the guarantee rests entirely on $\tau$'s recall. At SafetyMargin's 0.93 recall the residual miss rate is 0.07 (the nine SafetyMed items behind the 4.5-point gap of \S\ref{sec:exp:behavior}); tighter bounds need a higher-recall $\tau$ or an inference-time check. The framework already offers three recall-oriented mitigations: abstain-as-hard routing (Algorithm~\ref{alg:typecompact}), a lower promotion threshold $\theta_C$, and a one-time human audit of the flagged constraints. The guarantee covers the storage layer and does not address alignment-time training, chain-of-thought scaffolding, episodic consolidation, or inter-agent communication; items whose constraint status depends on session context need an online $\tau$.

\paragraph{Limitations of the evaluation.} \looseness=-1 AgentArtifactCorpus is drawn from public GitHub, so type distributions in closed enterprise corpora may differ; the released classifier should be re-fitted before deployment, with rubric, labels, and fitting code released for this. Annotators disagree at the five-class level ($\kappa = 0.45$, against 0.79 on the safety-critical split; \S\ref{sec:exp:classifier}), so per-type results away from the constraint boundary are correspondingly less reliable. Replication overhead under TypeDecompose varies widely (median 0\%, worst case 219\%; \S\ref{sec:exp:safety}); it is the cost of copying each constraint into every partition its scope covers. We ran the multi-round rollout on two of the four LLM families; single-round survival is a consistent 19--42\% at 50\% compression across all four. MaRS~\cite{mars_2025} has no public implementation, so Table~\ref{tab:curves} uses MaRS-FL, our reimplementation of its design (single submodular utility, no per-type pinning); a reference implementation may shift the numbers, but not the structural argument: single-utility selection cannot guarantee per-type bounds (\S\ref{sec:framework:operators}). 

\section{Conclusion}
\label{sec:conclusion}

Knowledge Triage assigns every item in an agent's knowledge base to one of five types and routes each through its own retention policy across compaction, decomposition, and retrieval. On five public corpora, the typed operators preserved safety constraints where uniform strategies lost them: 2--4$\times$ higher constraint recall in compaction (Table~\ref{tab:curves}), 0\% versus 93\% locality violations in decomposition (Table~\ref{tab:dec}), and 100\% versus 73\% recall@50 in retrieval (Table~\ref{tab:ret}). The same per-type retention shows up downstream: on SafetyMed ($N = 200$) TypeCompact outperforms the production Sonnet compactor on pass rate ($p = 0.022$) and constraint preservation ($14.5$-point lead, $p < 10^{-8}$); on $\tau$-bench retail ($N = 115$ tasks, 1{,}019 paired rollouts) it outperforms the full-policy baseline and hierarchical truncation on pass rate ($p = 0.003$, $p = 0.005$); on the held-out $\tau$-bench airline split it outperforms hierarchical truncation ($p = 0.024$) and statistically ties the full-policy ceiling, consistent with generalization across customer-service domains. We summarize the resulting design principle as follows: before delegating compaction to a frontier LLM, classify items by safety role and retain the safety-critical class without modification.

Bounded-context management has moved from research prototypes into production systems that draft medical notes, write legal arguments, and modify customer accounts; safety preservation at the storage layer is therefore a deployment requirement any production agent runtime has to meet. 





\bibliographystyle{ACM-Reference-Format}
\bibliography{references}

\end{document}